\documentclass{article}

\usepackage[final]{timeseries_workshop}

\usepackage[utf8]{inputenc}
\usepackage[T1]{fontenc}

\usepackage{amsmath,amssymb}
\usepackage{graphicx}
\usepackage{multirow}
\usepackage{subcaption}
\usepackage{xcolor}
\usepackage{booktabs}

\usepackage{hyperref}
\usepackage{url}

\usepackage[ruled,vlined]{algorithm2e}
\usepackage{float}
\usepackage{placeins} 
\usepackage{paralist}

\usepackage[capitalise,nameinlink]{cleveref}

\title{Adaptive Regime-Switching Forecasts with Distribution-Free Uncertainty: Deep Switching State-Space Models Meet Conformal Prediction
}

\definecolor{linkcolor}{RGB}{83,83,182}

\hypersetup
{
  colorlinks=true,
  citecolor=linkcolor,
  linkcolor= linkcolor,
  urlcolor=blue,
  pdfstartview=FitH,
  pdftitle={},
  pdfauthor={},
  pdfcreator={},
  pdfsubject={},
  pdfkeywords={},     
}

\newcommand{\dstm}{DS\textsuperscript{3}M}

\author{{Echo Diyun LU}\\
University of Lorraine, CNRS
\And 
Charles Findling\\
Pernod Ricard\\
\And 
Marianne Clausel\\
University of Lorraine
\And 
Alessandro Leite\\
INSA Rouen Normandy
\And
Wei Gong\\
Pernod Ricard
\And 
Pierric Kersaudy\\
Pernod Ricard
}

\begin{document}

\maketitle

\begin{abstract}
Regime transitions routinely break stationarity in time series, making calibrated uncertainty as important as point accuracy. We study distribution-free uncertainty for regime-switching forecasting by coupling Deep Switching State Space Models (\dstm) with Adaptive Conformal Inference (ACI) and its aggregated variant (AgACI). We also introduce a unified conformal wrapper that sits atop strong sequence baselines—including S4, MC-Dropout GRU, sparse Gaussian processes, and a change-point local model to produce online predictive bands with finite-sample marginal guarantees under non-stationarity and model misspecification. Across synthetic and real datasets, conformalized forecasters achieve near-nominal coverage with competitive accuracy and generally improved band efficiency.
\end{abstract}

\section{Introduction}\label{sec:intro}

Time-series forecasting often involves hidden regime switches, abrupt transitions between latent operating modes characterized by changes in level, trend, volatility, or seasonality. Such shifts induce distributional nonstationarity, making not only point prediction but especially uncertainty quantification (UQ) highly challenging. By UQ, we mean the construction of well-calibrated predictive intervals, rather than merely accurate mean forecasts, since reliable decisions depend on risk-aware intervals rather than single-value predictions.

\textbf{Why UQ under regime switching is difficult.}
When regimes change, predictive distributions can become severely miscalibrated: intervals may undercover or inflate excessively, even when point errors remain small. Classical UQ approaches fail to handle regime switches properly. For e.g., Gaussian processes with stationary kernels or Bayesian models with smooth priors implicitly assume differentiability or stationarity. These assumptions fail at regime boundaries, where the calibration typically fails. Thus, UQ in this setting requires methods that are both \emph{adaptive} and \emph{robust to nonstationarity}, capable of reacting online to structural shifts.%

\textbf{Our approach.}
We combine two complementary tools:  \begin{inparaenum}[(a)] \item \emph{Deep Switching State--Space Model} (DS$^3$M) that explicitly incorporates a discrete regime variable to capture mode transitions while retaining continuous latent dynamics for within-regime forecasts~\cite{xu2025deepswitchingstatespace}; and  \item \emph{Adaptive Conformal Inference} (ACI) and its aggregated variant (AgACI), which yield \emph{distribution-free predictive intervals} by correcting miscoverage online in dependent time series~\cite{zaffran2022adaptiveconformalpredictionstime}.\end{inparaenum}
This coupling achieves both explicit regime modeling and online calibration. In particular, DS$^3$M provides regime-aware point forecasts, while ACI/AgACI ensures valid coverage guarantees despite temporal dependence and non-i.i.d.\ residuals. Whereas vanilla conformal prediction relies on exchangeability and fails under regime switches, adaptive updates combined with regime-aware residuals make the wrapper effective in streaming, nonstationary settings. In summary, our contributions are as follows:

\phantom{~~~~}$\bullet$ {\bf A new agnostic calibration protocol:} We introduce a protocol-agnostic calibration layer for regime-switching forecasting by combining DS$^3$M with adaptive conformal methods, yielding reliable, distribution-free prediction intervals. \\
\phantom{~~~~}$\bullet$ {\bf A new conformal-based wrapper:} We propose a consistent residual-based conformal wrapper applicable across strong baselines (S4~\cite{gu2022efficientlymodelinglongsequences}, MC-Dropout GRU, CPD~\cite{TRUONG2020107299}, and Sparse GP), ensuring fair comparisons.\\
\phantom{~~~~}$\bullet$ {\bf Empirical validation in different datasets:} We conduct experiments on synthetic and real datasets exhibiting regime changes, reporting coverage at 90\% and median interval width. Experimental results demonstrate that our approach produces well-calibrated adaptive intervals that respond to regime shifts without sacrificing point accuracy.

\section{Related Work}\label{sec:related}

\textbf{Regime-switching state-space models.}
Classical switching state-space models (SSMs), as well as recent deep variants such as DS$^3$M, incorporate a discrete regime variable to capture transitions while maintaining continuous latent states for within-regime dynamics. Variational inference enables scaling to long sequences~\cite{gu2022parameterizationinitializationdiagonalstate}. These models offer robust regime-aware forecasting capabilities; however, their uncertainty quantification often relies on Bayesian approximations, which can be computationally expensive and prone to miscalibration following abrupt changes.

\textbf{Distribution-free uncertainty for time series.} Conformal Prediction (CP) offers distribution-free guarantees of marginal coverage under exchangeability. However, the i.i.d.\ assumption breaks in time-series settings. Adaptive Conformal Inference (ACI) addresses this by correcting online for temporal dependence and distributional shifts, while Aggregated ACI (AgACI) stabilizes updates by combining multiple experts~\cite{zaffran2022adaptiveconformalpredictionstime}. These methods yield valid coverage in streaming data, but they do not explicitly model regime structures, limiting their responsiveness to abrupt mode changes.

\textbf{Long-context state-space sequence models.}  
Recent structured state-space layers, such as S4, efficiently capture long-range dependencies and have become strong neural backbones for sequential modeling~\cite{gu2022efficientlymodelinglongsequences}. While highly effective for point forecasting, they do not inherently address the challenges of calibrated UQ under regime switching.

\textbf{Baselines for uncertainty.}  
Several families of methods have been applied to uncertainty in nonstationary sequences. Change-point detection (CPD) explicitly segments data around structural breaks~\cite{TRUONG2020107299}, Bayesian deep nets such as MC-Dropout GRU~\cite{pmlr-v48-gal16} provide scalable model-based uncertainty, and Gaussian processes (GPs) offer uncertainty quantification through kernels~\cite{li2025gaussianprocessesregressionuncertainty}. Yet each faces limitations: GP methods are unscalable and unadaptive; Bayesian deep nets lack guaranteed coverage and miscover after switches; and UQ in deep switching models often inherits the limitations of approximate Bayesian inference.
\textbf{Positioning.}  
Vanilla CP ensures distribution-free coverage but assumes exchangeability, which is violated under regime shifts. ACI relaxes this assumption through online calibration, while DS$^3$M directly models discrete regime transitions. Our work combines these complementary perspectives: by embedding adaptive conformalization within a regime-switching model, we obtain calibrated, distribution-free intervals that remain reliable across regime changes while retaining competitive point forecast accuracy.

\section{Method}\label{sec:method}

\paragraph{DS\textsuperscript{3}M forecaster.}
Let $y_t$ denote the target series. DS\textsuperscript{3}M posits discrete regimes $d_t \in \{1,\ldots,K\}$ (Markov), continuous states $z_t$, and a learned history summary $h_t$. Given observations up to $t$, we form a one-step mean prediction $\hat{y}_t$ and define residual scores $s_t = |y_t - \hat{y}_t|$.

\paragraph{Adaptive Conformal Inference (ACI).}

Given a calibration buffer, ACI produces intervals $\widehat{C}_t = [\hat{y}_t \pm \widehat{Q}_{1-\alpha_t}]$, where $\widehat{Q}$ is the empirical quantile of past scores. The miscoverage level is updated online:
\[
\alpha_{t+1} = \alpha_t + \gamma\!\left( \alpha - \mathbf{1}\{y_t \notin \widehat{C}_t\} \right),
\]
with nominal $\alpha=0.1$ and learning rate $\gamma>0$. The pseudo-code is described in~\cref{app:aci}. 

\paragraph{Aggregated ACI (AgACI).}
We maintain a small set of ACI experts with different $\gamma$ values and aggregate their quantiles (e.g., via fixed or exponentially weighted averaging). This reduces sensitivity to the choice of $\gamma$.


\paragraph{Unified CP wrapper for baselines.}
We apply the same residual-based CP layer to each forecaster:\\
  \phantom{~~~~}$\bullet$ \textbf{S4 (S4D backbone):} encoder $\rightarrow$ S4D blocks $\rightarrow$ decoder; long-range modeling with residual CP.\\
  \phantom{~~~~}$\bullet$ \textbf{CPD:} online change-point detection segments the stream; within the last segment, we fit a light GRU forecaster.\\
  \phantom{~~~~}$\bullet$ \textbf{MC-Dropout GRU:} GRU with dropout kept at test time (posterior samples are \emph{not} used for coverage); CP uses absolute residuals of the predictive mean.\\
  \phantom{~~~~}$\bullet$\textbf{Sparse GP:} inducing points (GPyTorch); we use the posterior mean for residuals; CP bands provide distribution-free coverage.\\

\section{Experimental Setup}\label{sec:setup}

\textbf{Datasets.} 
Our evaluation harnesses both synthetic and real-world datasets. Synthetic components include the toy autoregressive process endowed with articulated breaks and Lorenz-driven perturbations. Real-world configurations include Sleep Apnea (2Hz), US Unemployment (monthly). Please see~\cref{app:data} for more details.

\noindent\textbf{Protocol.} 
The protocol executes one-step rolling forecasts across held-out test windows, with iterative multi-step configurations permitted when meaningful. We standardize each target series using a standard scaler derived from the training split and invert transformations post-prediction. Lag length L is preselected as a hyperparameter, with defaults set at 48 for sub-hourly data and incrementally smaller for monthly horizons.

\noindent\textbf{Conformal settings.} 
Conformal scenarios examined here operate at a nominal confidence level of $1{-}\alpha = 0.9$. ACI employs factors $\gamma \in \{10^{-4},10^{-3},10^{-2}\}$; the AgACI approach aggregates diverse estimators to eliminate the need for manual parameter tuning. Calibration relies on the latest non-overlapping, autoregressive window that concludes prior to the target instance. CP intervals are centered on $\hat{y}_t$.

\noindent\textbf{Metrics.} 
Performance is quantified using root-mean-square error (RMSE) across the held-out window, along with empirical Coverage@90\% (fraction in band) and the median interval width. We monitor the incidence of degenerate or infinite bands(rare in our runs).

\noindent\textbf{Training details.} 
Training regimes differ by model family. For parametric architectures (DS\textsuperscript{3}M, S4, GRU), we adopt Adam optimizers, micro-batch training across lagged sequences, and impose an early stopping criterion based on validation RMSE; mixed-precision arithmetic is engaged wherever the hardware supports it. Sparse Gaussian Processes are discipled via a variational evidence lower bound, permitting learnable inducing point locations and subject to identical early stopping criteria on validation RMSE.

\section{Results}\label{sec:results}

\begin{figure}
\centering
\begin{subfigure}{0.48\linewidth}
  \includegraphics[width=\linewidth]{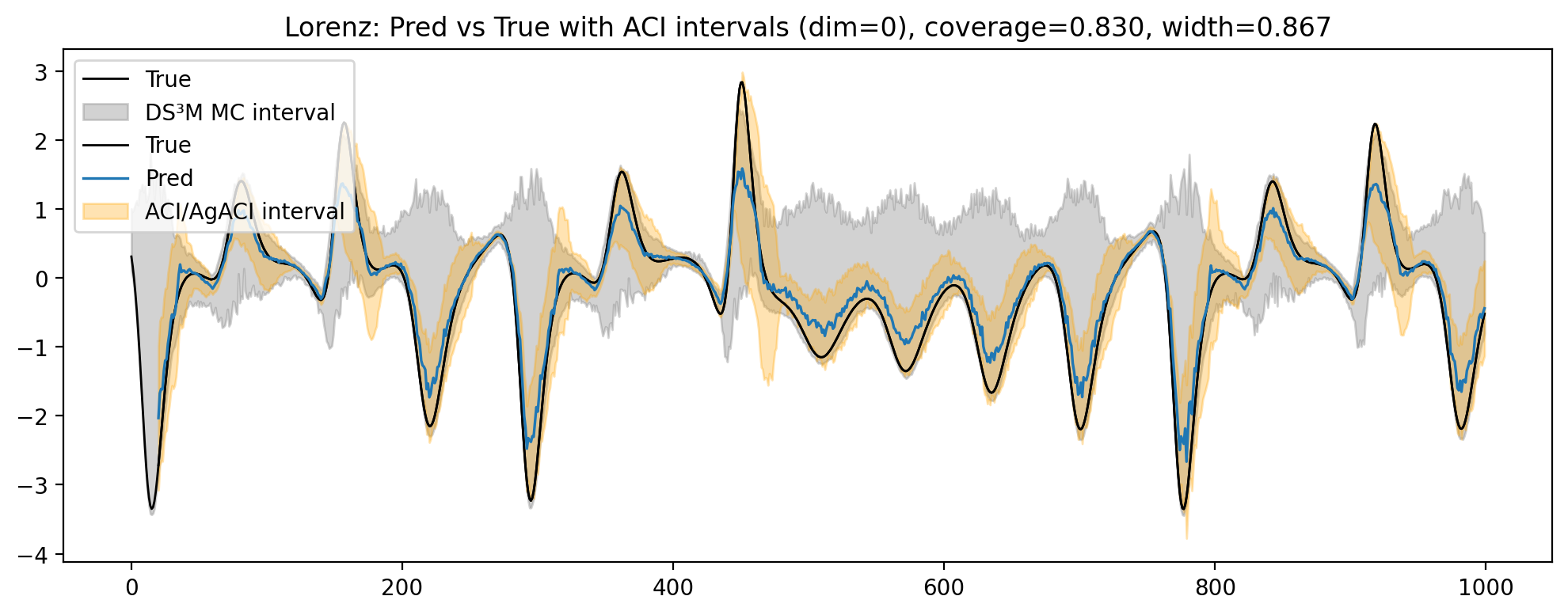}
  \caption{Lorenz: DS$^3$M + ACI}
\end{subfigure}\hfill
\begin{subfigure}{0.48\linewidth}
  \includegraphics[width=\linewidth]{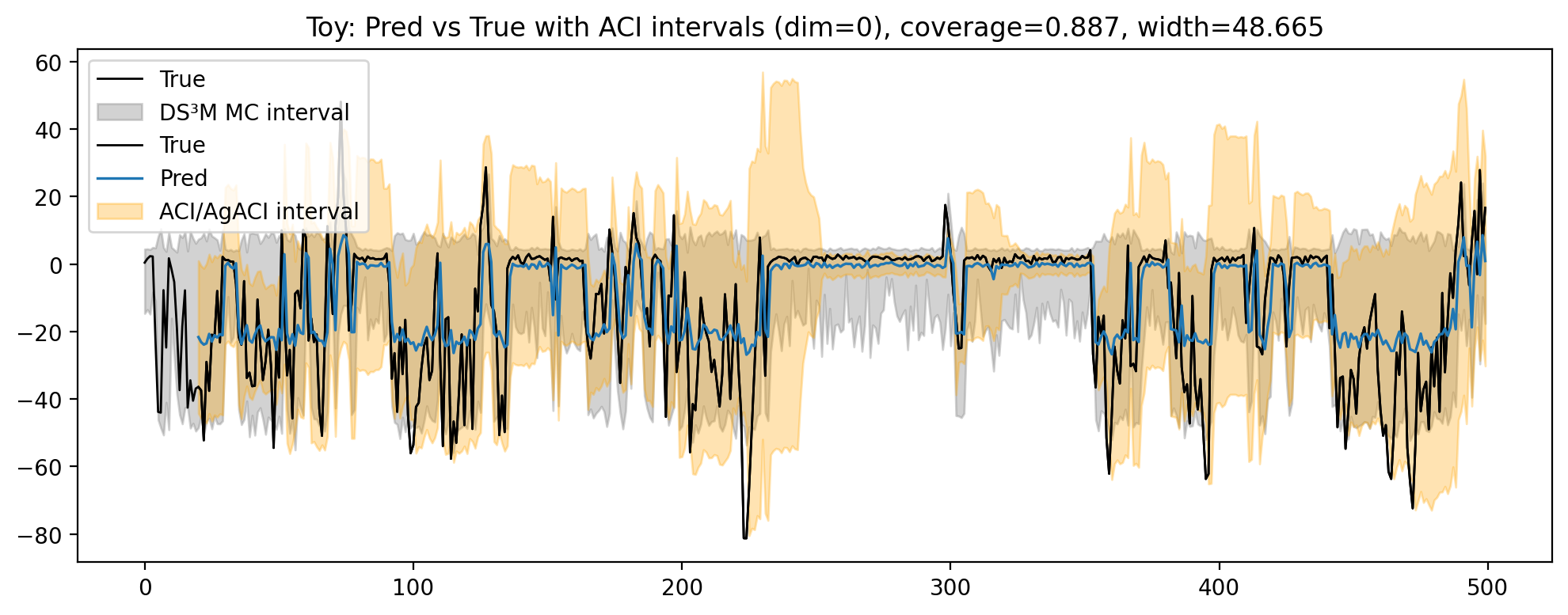}
  \caption{Toy: DS$^3$M + ACI}
\end{subfigure}

\begin{subfigure}{0.48\linewidth}
  \includegraphics[width=\linewidth]{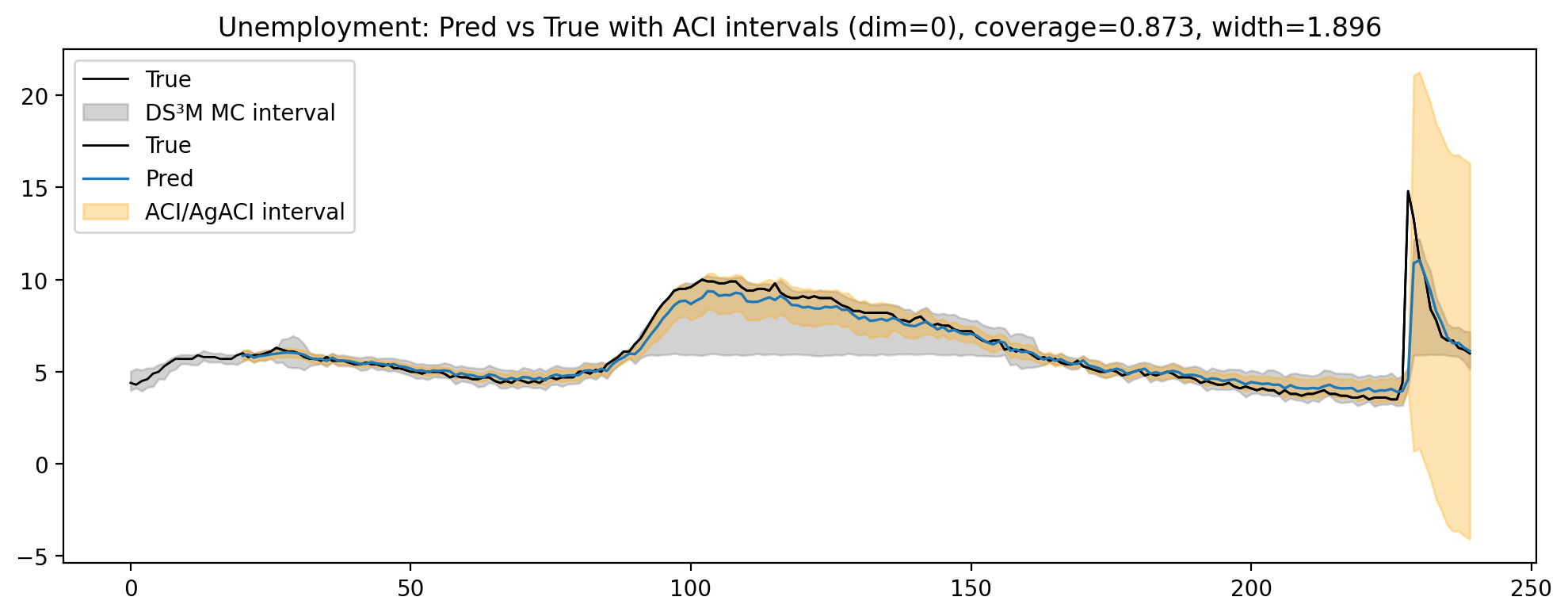}
  \caption{Unemployment: DS$^3$M + ACI}
\end{subfigure}\hfill
\begin{subfigure}{0.48\linewidth}
  \includegraphics[width=\linewidth]{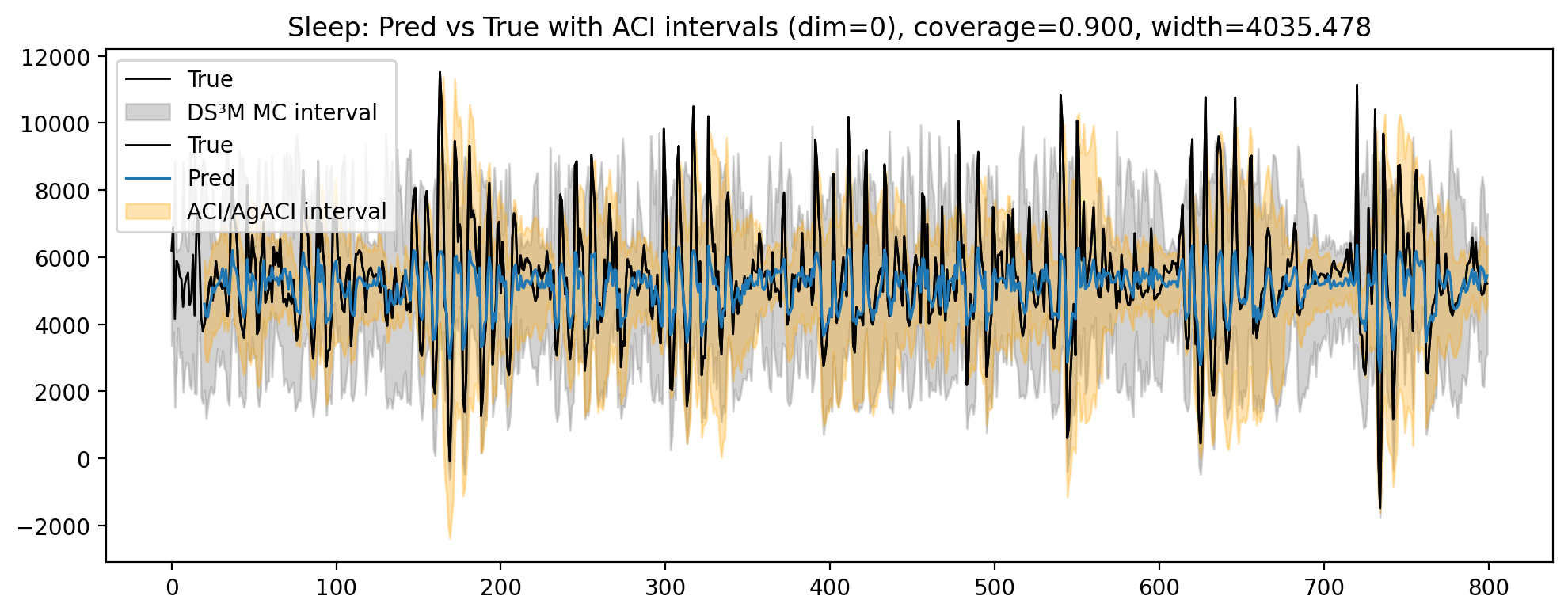}
  \caption{Sleep: DS$^3$M + ACI}
\end{subfigure}

\caption{\textbf{Prediction vs.\ truth with ACI intervals.} Black: observed series; Blue: DS$^3$M mean; Grey: DS$^3$M native MC interval; Orange: ACI band. Coverage near $0.90$ indicates successful calibration; narrower orange bands at similar coverage indicate more efficient uncertainty than the native MC band.}
\label{fig:aci-results}
\end{figure}

As shown in Figure~\ref{fig:aci-results}: the orange bands of adaptive conformal inference are distribution-free and adapt online. When the gray MC band is too wide or too narrow, the conformal band corrects for miscalibration. Dataset-to-dataset differences show how regime complexity and periodicity shape both residuals and interval efficiency.

Conformalization preserves all models relatively close to the 90\% target across datasets, but the regime also DS$^3$M with AgACI achieves the best calibration overall--giving the highest coverage on Lorenz, Unemployment, and Sleep datasets. For efficiency, DS$^3$M generates the narrowest bands for large-scale Sleep series, and the S4 backbone has the lowest interval on the Lorenz dataset. In summary, DS$^3$M combined with AgACI yields strong calibration on the harder, regime-heavy set, while S4 does best interval-wise on the smoother, low-noise string. 

\vspace{-1em}

\begin{table}[H]
\centering
\small
\caption{Comparison of Coverage and Median Interval Length on different datasets.}
\label{tab:cov_len_block}
\setlength{\tabcolsep}{3pt}
\begin{tabular}{l *{5}{r} *{5}{r}}
\toprule
& \multicolumn{5}{c}{\textbf{Coverage@90\%}} & \multicolumn{5}{c}{\textbf{Median Interval Length}} \\
\cmidrule(lr){2-6} \cmidrule(lr){7-11}
\textbf{Datasets} &
\textbf{CPD} & \textbf{GP} & \textbf{MCD} & \textbf{S4} & \textbf{DS$^3$M} &
\textbf{CPD} & \textbf{GP} & \textbf{MCD} & \textbf{S4} & \textbf{DS$^3$M} \\
\midrule
Lorenz        & 0.863 & 0.893 & 0.909 & 0.904 & \textbf{0.910} & 0.199 & 0.107 & 0.122 & 0.032 & \textbf{0.127} \\
Unemployment  & 0.858 & 0.888 & 0.846 & 0.862 & \textbf{0.893} & 23.860 & 2.196 & 0.570 & 0.558 & \textbf{0.728} \\
Sleep         & 0.901 & 0.898 & 0.901 & 0.900 & \textbf{0.904} & 4041.667 & 3915.016 & 3717.248 & 4297.769 & \textbf{3507.418} \\
\bottomrule
\end{tabular}
\end{table}

\vspace{-1em}

\FloatBarrier 

\section{Discussion and Future Direction}
\label{sec:discussion}
A central finding is that adaptive conformal wrappers provide reliable, distribution-free uncertainty estimates that are largely independent of the underlying forecasting architecture. For S4, GRU, and SVGP, conformalization aligns coverage with targets while maintaining accuracy and interval efficiency, making it a practical calibration layer in the presence of distribution shift. DS$^3$M’s regime-aware structure aids interpretability and point prediction, yet its native uncertainty can be fragile under sharp transitions and heavy-tailed residuals; residual-based conformalization corrects coverage but may still inherit scale sensitivity in challenging regimes. Overall, calibrating via residual scores with online adaptation offers a low-friction path to trustworthy intervals without retraining, complementing both classical and deep modeling approaches and aligning with the broader goal of robust, foundation-model-ready time-series evaluation.

The coverage guarantees offered here are marginal rather than conditional, meaning that they do not hold within specific regimes or covariate groups, which may be important for high-stakes applications. Using absolute residuals yields symmetric bands, which can be inefficient under skewed or heteroskedastic noise, and the calibration buffer introduces a short delay before intervals adapt after sudden shifts. Comparing median widths across datasets of very different scales can also distort efficiency assessments, sometimes exaggerating interval size for scale-sensitive models. Finally, our evaluation focuses on one-step predictions and does not fully examine the quality of native probabilistic forecasts or decision-oriented performance under asymmetric costs. Future work will explore regime-aware conformal methods that explicitly condition on inferred regimes to improve coverage within each regime and tighten intervals. 

\section*{Acknowledgments}

This research was partially funded by the French National Research Agency (ANR) under Grants No. 23-PEIA-004 and No. ANR-23-CPJ1-0099-01.

\bibliographystyle{abbrvnat}
\bibliography{references}

\appendix
\section{Conformal Prediction Primer}
\label{app:cp}
Conformal prediction (CP) turns any point forecaster into a distribution-free interval predictor with finite-sample \emph{marginal} coverage. Given a forecaster producing one-step means $\hat{y}_t$ and residual scores $s_t=\lvert y_t-\hat{y}_t\rvert$, a split-CP interval at time $t$ is
\[
\widehat{C}_t=\big[\hat{y}_t \pm \widehat{Q}_{1-\alpha}\big],
\]
where $\widehat{Q}_{1-\alpha}$ is the empirical $(1-\alpha)$-quantile of calibration scores. Under exchangeability this ensures $\Pr\{y_t\in \widehat{C}_t\}\approx 1-\alpha$.
In time series, exchangeability is violated; \emph{Adaptive Conformal Inference} (ACI) compensates by updating the target miscoverage online:
\[
\alpha_{t+1}=\alpha_t + \gamma\big(\alpha-\mathbf{1}\{y_t\notin \widehat{C}_t\}\big),
\]
So intervals widen after misses and narrow after hits. \emph{AgACI} aggregates multiple ACI experts with distinct learning rates $\gamma$, stabilizing adaptation under heterogeneous conditions. In this work, the CP layer is \emph{residual-based and model-agnostic}; we center at $\hat{y}_t$ and use absolute residuals by default, with studentized/variance-aware scores left for future work. See Algorithm~\ref{app:aci} for pseudo-code.


\section{What We Mean by Regimes and Switching}
\label{app:regime}
A \emph{regime} is a latent operating mode with distinct dynamics (level, trend, variance, seasonality). We write $d_t\in\{1,\dots,K\}$ for the discrete regime (often Markov), a continuous state $z_t$, and observation $y_t$, e.g.,
\[
z_{t+1}=f_{d_t}(z_t)+\eta_t,\qquad y_t=g_{d_t}(z_t)+\epsilon_t.
\]
Switches in $d_t$ produce non-stationarity (abrupt changes in mean/variance/correlation). Unlike Gaussian Processes or Bayesian NNs that typically encode non-stationarity via kernels or parameter uncertainty, DS$^3$M \emph{explicitly} models discrete mode changes while keeping continuous dynamics, which improves interpretability and one-step accuracy. Change-point detection (CPD) instead \emph{segments} the series and fits local models on the last segment. Our conformal layer sits on top of either approach and adapts online after switches (Appendix~\ref{app:cp}).

\section{Datasets and Summary}\label{app:data}

\Cref{tab:datasets} summarizes the used datasets. We standardize each series using training statistics and invert the scale post-prediction.  

\begin{table}[H]
\centering
\small
\caption{Description of the datasets. $D$ is the dimensionality of the target series, $T$ the total number of datapoints.}
\label{tab:datasets}
\begin{tabular}{lccc}
\toprule
\textbf{Dataset} & \textbf{Frequency} & $D$ & $T$ \\
\midrule
Toy            & synthetic   & 1    & 3000 \\
Lorenz         & synthetic   & 1    & 10000 \\
Sleep Apnea    & 0.5 sec     & 1    & 2000 \\
Unemployment   & month       & 1    & 879  \\
\bottomrule
\end{tabular}
\end{table}

\section{ACI Pseudo-code}\label{app:aci}
\begin{algorithm}[H]
\DontPrintSemicolon
\SetAlgoLined
\caption{Adaptive Conformal Inference (ACI) around a one-step forecaster}
\KwIn{Forecaster $\hat{y}_t$, nominal $\alpha$, learning rate $\gamma$, initial $\alpha_{t_0}$, residual buffer $\mathcal{S}_{t_0}$.}
\For{$t = t_0, t_0{+}1, \dots$}{
  Form $\mathcal{S}_t$ from absolute residuals up to $t{-}1$.\;
  Compute $\widehat{Q}_{1-\alpha_t}$, the $(1{-}\alpha_t)$ empirical quantile.\;
  Predict $\widehat{C}_t = [\hat{y}_t \pm \widehat{Q}_{1-\alpha_t}]$.\;
  Observe $y_t$ and update $\alpha_{t+1} = \alpha_t + \gamma(\alpha - \mathbf{1}\{y_t \notin \widehat{C}_t\})$.\;
}
\end{algorithm}

\end{document}